\documentclass[letterpaper, 10 pt, conference]{ieeeconf}  

\IEEEoverridecommandlockouts                              

\title{\LARGE \bf
{Attentional Network for Visual Object Detection}
}

\author{Kota Hara\thanks{This work was done while the first author was an intern at MERL} \\
University of Maryland, College Park\\
{\tt\small kotahara@terpmail.umd.edu}
\and
Ming-Yu Liu, Oncel Tuzel, Amir-massoud Farahmand\\
Mitsubishi Electric Research Laboratories\\
{\tt\small \{mliu,oncel,farahmand\}@merl.com}
}

\usepackage{algorithm}
\usepackage{algorithmic}
\usepackage{booktabs} 
\usepackage{multirow}
\usepackage{amsmath}
\usepackage{amssymb}
\usepackage{amsfonts}
\usepackage{graphicx}

\newcounter{assumption}
\renewcommand{\theassumption}{A\arabic{assumption}}

\renewcommand{\AA}{{\mathcal{A}}}

\newcommand{\XX}{{\mathcal{X}}}

\newcommand{\beq}{\begin{equation}}
\newcommand{\eeq}{\end{equation}}
\newcommand{\beqa}{\begin{eqnarray}}
\newcommand{\eeqa}{\end{eqnarray}}
\newcommand{\beqan}{\begin{eqnarray*}}
\newcommand{\eeqan}{\end{eqnarray*}}
\newcommand{\ben}{\begin{eqnarray*}}
\newcommand{\een}{\end{eqnarray*}}

\newcommand{\EE}[1]{{\mathbb E}\left[#1\right]}

\newcommand{\eqdef}{\triangleq}

\newcommand{\EEX}[2]{{\mathbb E}_{#1}\left[#2\right]}

\newcommand{\PKernel}{\mathcal{P}}
\newcommand{\RKernel}{\mathcal{R}}

\begin{document}

\maketitle
\thispagestyle{empty}
\pagestyle{empty}

\begin{abstract} 
We propose augmenting deep neural networks with an attention mechanism for the visual object detection task. As perceiving a scene, humans have the capability of multiple fixation points, each attended to scene content at different locations and scales. However, such a mechanism is missing in the current state-of-the-art visual object detection methods. Inspired by the human vision system, we propose a novel deep network architecture that imitates this attention mechanism. As detecting objects in an image, the network adaptively places a sequence of glimpses of different shapes at different locations in the image. Evidences of the presence of an object and its location are extracted from these glimpses, which are then fused for estimating the object class and bounding box coordinates. Due to lacks of ground truth annotations of the visual attention mechanism, we train our network using a reinforcement learning algorithm with policy gradients. Experiment results on standard object detection benchmarks show that the proposed network consistently outperforms the baseline networks that does not model the attention mechanism.
\end{abstract} 

\section{Introduction}
\label{introduction}

Object detection is one of the most fundamental problems in computer vision. The goal of object detection is to detect and localize all instances of pre-defined object classes in the image, typically in the form of bounding boxes with confidence values. Although an object detection problem can be converted to many object classification problems by the scanning window technique~\cite{ViolaJones,liu2016unsupervised}, it is inefficient since a classifier has to be applied to all hypothesized image regions that are of various locations, scales, and aspect ratios. Recently, the region-based convolution neural network (R-CNN)~\cite{RCNN} algorithm adopts a two-stage approach. It first generates a set of object proposals, called regions of interest (ROI), using a proposal generator and then determines the existence of an object and its classes in the ROI using a deep neural network. The R-CNN algorithm achieves impressive performance on public benchmarks and has become the backbone of many recent object detection methods.

\begin{figure*}[hbt!]
  \centering
  \includegraphics[width=0.92\textwidth,natwidth=4550,natheight=700]{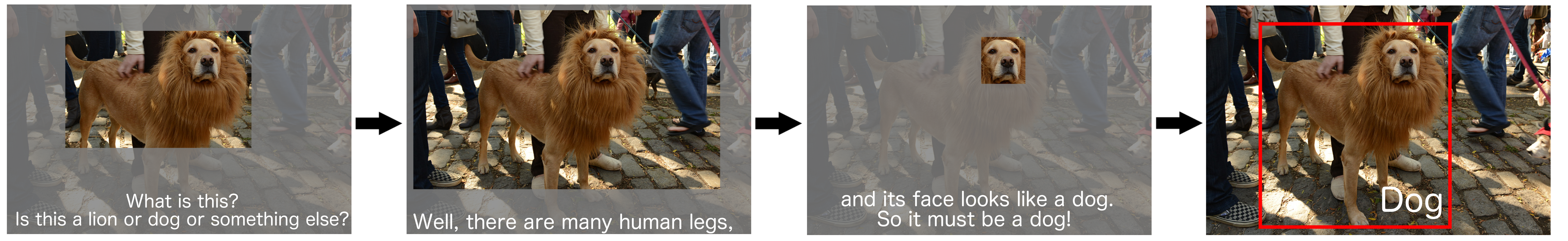}
    \caption{Humans have the capability of  analyzing image content for visual object detection from multiple fixation points.}
  \label{fig:illustration}  
    \includegraphics[width=0.92\textwidth,natwidth=847,natheight=239]{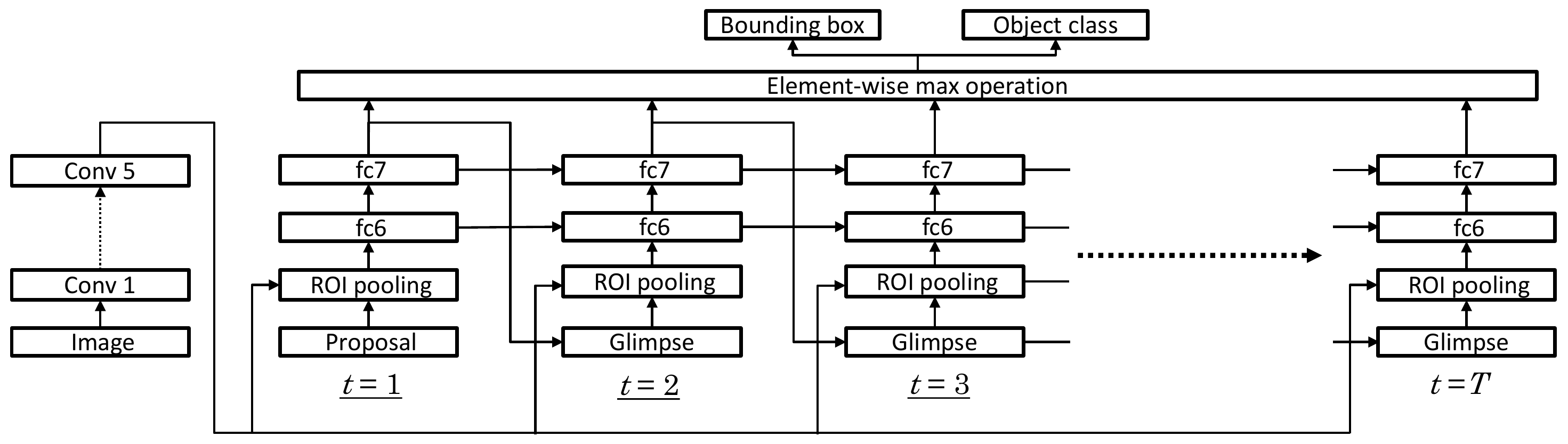}
  \caption{\small Illustration of the AOD network: the network consists a stacked recurrent module designed for object class recognition, bounding box regression and glimpse generation. The classification and bounding box regression are done only at the final time step while the glimpse generation is done at all time steps except the last time step. Given an input image, first, a set of feature maps are computed by the Deep Convolutional Neural Network. Given a proposal bounding box at $t=1$, a fixed dimensional feature vector is extracted from the region of the last feature map within the proposal bounding box using the ROI pooling layer~\cite{FastRCNN}. A few fully connected layers (fc6 and fc7 in the figure), each followed by a ReLU and dropout layers, are then applied to the extracted feature vector. From the resultant features, a next glimpse bounding box is determined by applying a fully connected layer. At $t=2$, a feature vector is extracted again by the ROI pooling layer. The process repeats until the last time step $t=T$. At the last time step, an element-wise max operation is applied to the final feature vectors at all time steps and then softmax classification and bounding box regression are conducted.}
  \label{fig:flowchart_detailed}
\end{figure*}

As detecting an object, the R-CNN algorithm and its extensions look at the proposal region (and sometimes its neighborhood) given by the proposal generator only once. This is in contrast to humans' capability of multiple fixations of visual attention as depicted in Fig.~\ref{fig:illustration}. We propose to imitate such an attention mechanism for improving the object detection performance of the R-CNN algorithm. To this end, we design a novel deep network architecture that adaptively places a sequence of glimpses for accumulating visual evidence for determining the object class and its precise location from a proposal region. We use a recurrent neural network for determining the glimpse placements (locations and sizes) as well as for summarizing the visual evidence extracted from the glimpses. Due to lacks of ground truth annotations of the visual attention mechanism for the object detection task, we train the proposed network using a reinforcement learning algorithm.

Our work is largely inspired by~\cite{mnih2014recurrent}, which propose a visual attention mechanism for the handwritten digit classification task. Our work is different in that we study the application of the attention mechanism for the more challenging object detection task. Due to the large variation of visual objects in appearances deformations and scales, it is more difficult to learn a reliable attention mechanism. The glimpse has to vary both in shapes and scales for finding more relevant information. We investigate into various network design choices as well as training methods for finding the network structure that can facilitate the learning of a reliable visual attention mechanism for the object detection task. We provide detailed performance analysis on these choices in the experiment section. We evaluate the proposed algorithm, which we refer to as the Attentional Object Detection (AOD) Network, using the PASCAL VOC detection benchmarks. Through consistent performance boost over the baseline R-CNN algorithm under various settings, we conclude the advantage of utilizing the attention mechanism for visual object detection.

\section{Related Work}
\label{relatedwork}

The attention mechanism has been proposed for different applications including speech recognition~\cite{Chorowski2015}, machine translation~\cite{Bahdanau2015} and question--answering~\cite{Sukhbaatar2015}. Particularly, \cite{mnih2014recurrent} propose a recurrent neural network that sequentially selects image regions and combines extracted information from these regions for the handwritten digit classification task.

In this paper, we extend~\cite{mnih2014recurrent} to deal with the visual object detection task, which is considered a much more difficult task due to various appearance variations visual objects can exhibit in images. Moreover, unlike the classification task, the visual object detection task also requires an algorithm to localize the objects from various classes present in an image. While the glimpse size and shape is fixed in~\cite{mnih2014recurrent}, the glimpse size and shape is adaptively changed for maximizing the object detection performance.

A few attention-based methods are proposed for the object detection task. \cite{Caicedo2015}~train a class specific object localization model using a reinforcement learning algorithm and utilize the model for a detection task by evaluating all the regions generated over the course of localization. \cite{Yoo2015}~propose a class specific model which iteratively modifies the initial ROI until it declares the existence of an object. Unlike these works, the proposed method is class agnostic, which scales better as dealing with large amount of object classes. We do not need to train a detector for each object class. 

Most of the recent object detection methods are based on the R-CNN algorithm and explore two directions for further performance improvement. The first direction is to make the underlying CNN deeper \cite{arXiv:1512.03385}. The second one is to incorporate semantic segmentation \cite{arXiv:1512.04412}, which typically require additional training data for semantic segmentation. Other works focus on speeding up the computation time  \cite{FastRCNN,FasterRCNN,arXiv:1506.02640,arXiv:1506.06981,arXiv:1512.07729}.

An attempt to extract features from multiple regions is made by a few works. In \cite{MR-CNN}, in addition to the proposal bounding box, visual features are extracted from a set of hand-chosen regions. In \cite{chen2016small}, an additional context region is used for detecting small objects in images. The work of~\cite{arXiv:1512.04143} extracts features from the entire image in addition to the proposal regions for incorporating the context information. In contrary, the proposed approach does not rely on manual region selection. It learns to adaptively select regions based on the image content.

\section{Attentional Visual Object Detection Network}
\label{method}

We describe the AOD network in details. The network is a deep recurrent neural network designed to detect objects in an image by placing a sequence of glimpses of different sizes and aspect ratios and make a final decision based on features extracted from these glimpses. Each sequence starts from an object proposal bounding box given by the proposal generator and at the end of the sequence, the network produces scores and bounding boxes for all of the pre-defined classes. With the help of a reinforcement learning algorithm, the network is trained to generate glimpses that lead to better detection performance. In the following, we first describe our network behavior in the testing time, then we briefly introduce the reinforcement algorithm and the training of the proposed network.

\subsection{Network Architecture}

The AOD network as illustrated in Fig.~\ref{fig:flowchart_detailed} is an active recurrent neural network that decides the attention areas by itself. Given an image, the detection process starts by first applying a deep convolutional neural network to the whole image to obtain a set of feature maps as in the Fast R-CNN algorithm~\cite{FastRCNN}. In the case of utilizing pre-trained networks such as the AlexNet or the VGGNet, the feature maps are computed from the last convolutional layers. Concurrently, a set of proposal bounding boxes are obtained by running a proposal generator. The AOD processes each proposal bounding box separately by extracting the features from the computed feature maps within the bounding box regions. In the following, we describe a procedure applied to each proposal bounding box.

We denote a glimpse at each time step $t$ by $G_{t} \in \mathbb{R}^4$. The first glimpse, $G_{1}$, is a proposal bounding box given by the proposal generator and the subsequent $G_{t}$ are dynamically determined by the network by aggregating information acquired so far. As in \cite{RCNN}, we employ the scale-invariant and height/width normalized shift parameterization for $G_{t}$ by using the proposal bounding box as an anchor bounding box. Specifically, 
\begin{equation}
G = (\delta_x,\delta_y,\delta_w,\delta_h)=( \frac{ g_x - p_x }{ p_w},\frac{ g_y - p_y }{ p_h}, \log \frac{g_w}{p_w} , \log \frac{g_h}{p_h} )\nonumber
\end{equation}
where $(g_x, g_y, g_w, g_h)$ is the center coordinate, width and height of the glimpse bounding box. Similarly, $(p_x, p_y, p_w, p_h)$ represents the proposal bounding box. The glimpse layer generates $(\delta_x,\delta_y,\delta_w,\delta_h)$ for determining the glimpse bounding box, which is considered as the next glimpse for information aggregation. We note that the glimpse bounding boxes are not necessarily the object bounding boxes.

From each $G_{t}$, a fixed dimensional feature vector is extracted by applying the ROI pooling~\cite{FastRCNN} to the computed feature maps region within $G_t$. The ROI pooling operation divides a given ROI into a predefined grid of sub-windows and then max-pools the feature values in each sub-window. The pooled features are fed into a stack recurrent neural network of two layers, which are termed as fc6 and fc7 respectively.

At the last time step $t=T$, an element-wise max operation is applied to the last feature vectors at all time steps to compute the final feature vector. The final feature vector is fed into a softmax classification layer and bounding box regression layer for computing the object class and the object location. The softmax classification layer outputs class probabilities over $K$ foreground object classes and one background class. The bounding box prediction layer outputs bounding box prediction for each of the $K$ foreground classes. 

We note that the element-wise max operation retains the most strong signal across time steps independent of the order of the time steps. The stack recurrent network allows alternative paths of information propagation. They are used because of empirical evidence of superior performance as will be discussed in the experiment section.

\subsection{Reinforcement learning}
\label{sec:ReinforcementLearning}

The glimpse generation problem can be seen as a reinforcement learning (RL) problem~\cite{Sutton98,SzepesvariBook10}.
In RL, an agent continually interacts with an environment by observing the state $x \in \XX$ of the environment and then choosing an action $a \in \AA$ according to its policy $\pi(a|x)$, a probabilistic mapping from the state to actions. Depending on the current state and the chosen action, the agent's state in the environment changes to $X' \sim \PKernel(\cdot|x,a)$. The agent also receives a real-valued reward signal $r \sim \RKernel(\cdot|x,a)$.

This interaction might continue for a finite or infinite number of steps. In this paper, we consider a finite number of steps $T$. The outcome of each $T$ step of interactions is called an episode, which we denote by $\xi$.

The goal of an RL agent is to maximize the sum of the rewards it receives in the episode,  $R(\xi) = \sum_{t=1}^{T} r_t$, where $R(\xi)$ is the \emph{return} of $\xi$, and the goal of an RL can be stated as finding a policy $\pi$ so that the expected return $J(\pi) \eqdef \EEX{\pi}{R(\xi)}$ is maximized.

What differentiates RL from supervised learning is that there is no training data consisting of correct input-output pairs. Instead, the policy should be learned based only on the reward signal that the agent receives at each time steps. This is very appropriate for our problem as there is no dataset providing us with proper glimpse locations, but on the other hand it is relatively easy to specify whether the new glimpse location is useful for the task of object detection or not.

Among many different approaches to solve an RL problem, in this paper we use the REINFORCE algorithm~[\cite{Williams1992}], which is a policy gradient approach~\cite{DeisenrothNeumannPeters2013,SuttonMcAleesterSinghMansour2000}.
Suppose $\pi$ is parameterized by $\theta$. 
The policy gradient algorithm, in its simplest form, changes the policy parameters in the direction of gradient of $J(\pi_\theta)$ by the gradient ascent update, $\theta_{i+1} \leftarrow \theta_{i} + \alpha_i \nabla J(\pi_{\theta_i})$ for some choice of step size $\alpha_i > 0$.

By using the Gaussian distribution as $\pi_{\theta}$, the approximate gradients are computed by generating multiple episodes under the current policy (refer to \cite{Williams1992} for the derivation):
\begin{align}
\label{eq:EmpiricalPolicyGradient}
	\nabla_\theta J(\pi_{\theta})
	\approx
	\frac{1}{n}
	\sum_{i=1}^n
		R(\xi^{(i)}) 
		\sum_{t=1}^{T} \frac{(a_t^{(i)} - \theta x_t^{(i)}) x_t^{(i)\top}}{\sigma^2}.
\end{align}

Since this is a gradient ascent algorithm, it can easily be incorporated into the standard back propagation neural network training. In fact, our network is trained by back propagating both gradient from reinforcement learning and gradients from supervised training.

\subsection{Network Training}
\label{sec:training}

The training data fed to our network is constructed in the same way as that in the R-CNN algorithm. Each generated proposal bounding box is assigned a class label $c^*$ among one background class and $K$ foreground object classes according to the overlaps with the ground-truth object bounding boxes. The background class is anything not belonging to any of the foreground classes. Also given to each of the proposal bounding box is a bounding box target vector encoding the scale-invariant translation and log-space height/width shift relative to the object proposal as in the R-CNN method. Note that the bounding box target vector for background proposal boxes are not defined and thus not used for training.

The final outputs from our network are softmax classification scores and bounding boxes for all of the predefined foreground classes. During training, ground-truth annotations for them are provided, thus the standard Back Propagation Through Time (BPTT) algorithm \cite{58337} can be used for training. However, since the locations and shapes of the glimpses which lead to a higher detection performance are unknown, the BPTT algorithm cannot be applied to train the glimpse generation layer (an arrow from fc7 to Glimpse in the figure). The training of the glimpse generation layer is through the policy gradients described in Sec.~\ref{sec:ReinforcementLearning}. 

In our model, the state is the input given to the glimpse module (i.e., the output of fc7 in Figure~\ref{fig:flowchart_detailed}); the action is a new glimpse region described by a $G_t$ at time $t$. During training, we generate multiple episodes from each sample (a proposal bounding box). All episodes start from the same proposal bounding box and at each time step, i.i.d. Gaussian noise is added to the current glimpse region computed by the glimpse generation layer. For each episode, the network outputs class probabilities and object bounding boxes at the last time step. From these outputs, we compute a reinforcement reward for each episode as follows:
\begin{equation}\label{eq:return}
  r_t = \begin{cases}
    \mathrm{P}(c^*) \times \mathrm{IoU}( B_{c^{*}}, B_{c^{*}}^{*} ) & (t=T) \\
    0 & (otherwise)
  \end{cases}
\end{equation}
where $\mathrm{P}(c^*)$ is the predicted probability of the true class $c^*$ and $\mathrm{IoU}$ is the intersection over union between the predicted bounding box for $c^*$ and the ground-truth bounding box. Intuitively, if the glimpse bounding box after adding a Gaussian noise leads to a higher class probability and a larger IoU, then a higher return is assigned to the corresponding episode. The REINFORCE algorithm updates the model such that the generated glimpses lead to higher returns. In~\cite{mnih2014recurrent}, a 0-1 reward based on the classification success is used. We also evaluate a similar 0-1 reward and found that the proposed continuous reward function performs better for the object detection problem.

The AOD network is trained end-to-end by back propagating an expected gradient of the return along with other gradients computed from the standard classification and bounding box regression losses. The gradients from the REINFORCE algorithm affect all network parameters except those in the classification and bounding box regression layers. The gradients from the classification and localization layers affect all the parameters except those in the glimpse generation layer. We use the stochastic gradient descent with  mini-batches. 

The policy gradients are only computed for foreground samples because the appearance variations of the background class is significantly larger than those of the foreground classes and it is difficult for a reinforcement agent to learn a good glimpse generation policy. The net effect is that the glimpse generation is optimized only for better discrimination among foreground objects and more accurate bounding box regression. The benefit of excluding background samples for the REINFORCE algorithm is evaluated in Sec.~\ref{sec::design}.

\subsubsection{Return Normalization}

In the REINFORCE algorithm, typically a baseline is subtracted from the return in order to reduce the variance of the expected gradient while keeping the expected gradient unbiased. A common approach to obtain the baseline is to use exponential moving average of the return before subtracting the baseline~\cite{Williams1992}. Another approach is to learn a value function $V(x_t)=\EE{ \sum_{l=t}^{T} r_l | x_t }$ and use it as a baseline. 

We find out that in our setting, computing reliable baselines is challenging. The main reason is that our environment is a space of natural images whose variations are very large, and the agent is placed into a variety of different image sub-regions with different level of difficulties for making accurate decisions. Therefore, it is possible that all the episodes generated from a proposal bounding box $A$ get higher returns than those generated from a proposal bounding box $B$. In this case, all the episodes from $A$ are prioritized than those from $B$, which leads to an undesirable training behavior. 

To deal with this problem, we convert the original return in Eq.~\eqref{eq:return} by making the mean and variance of the returns computed from all episodes generated from one sample to 0 and 1, respectively, and use the converted return in the REINFORCE algorithm. This way, the new return reflects how well a particular episode works compared to other episodes from the same sample. Also the new return value is less dependent from the samples since it is normalized per sample. We find this approach works well in practice (Sec.~\ref{sec::design}). Note that the proposed return normalization scheme keeps the expected gradients unbiased as the computed baseline is the expectation over the rewards, which becomes a constant as computing the expected gradient.

\subsection{Implementation Details}
In this section, we present some of the implementation details which we omit from the main manuscript.

\subsubsection{Glimpse features}
At each time step, visual features are computed by the ROI pooling based on the glimpse generated in the previous time step. In addition to the visual features, we use the glimpse vector as an additional feature for the current time step fed into the stack recurrent neural network. This is to ensure that the network explicitly knows the glimpses it has produced. One fully connected layer followed by ReLU is applied to the glimpse vector and concatenated with the last visual feature vector (i.e., fc7 in Fig.~2 of the main manuscript). Similarly to fc6 and fc7, a recurrent connection is applied. Note that for $t=1$, the zero vector is fed as glimpse features. 

\subsubsection{Mini-batch training}
To reduce the memory footprint, one mini-batch contains samples from only a few images. Since the number of proposal boxes generated by a proposal generator such as the selective search algorithm from a single image is large, only a predefined number of foreground samples and background samples are randomly selected and used for training.

\subsubsection{Training sample construction}
\label{sec:sample_creation}
The training sample construction follows the procedure described in the Fast R-CNN work. For each sample, i.e., proposal bounding box $B$, we compute the IoU with all the ground-truth bounding boxes and select one with the highest IoU. Let $\alpha$ denote the highest IoU and $c$ denote the class label of the selected ground-truth bounding box. If $\alpha \geq 0.5$, we assign $c$ to $B$ and if $0.5 > \alpha \geq 0.1$, we assign the background class label to $B$. We ignore all other proposal bounding boxes for training. The whole process is done once before the start of the training stage.

\subsubsection{SGD hyper parameters}
\label{sec:sgd_hyper_parameters}
For each mini-batch, we randomly pick two training images and from each image, we randomly select 16 foreground samples and 48 background samples, resulting in 128 samples in one mini-batch. The glimpse generation layer is initialized from zero-mean Gaussian distributions with standard deviations 0.0001. The glimpse generation layer does not have a bias term. All the recurrent layers are initialized from zero-mean Gaussian with standard deviations 0.01 and the biases are set to 0. The fully connected layer applied to the glimpse vectors have 32 output neurons. We multiply the return by 0.1 to control the balance against the classification loss and regression loss. 


\subsubsection{Underlying Convolutional Network}
Our AOD uses a deep convolutional network (DCNN) to convert an input image into a set of feature maps. We evaluate AOD with two renowned DCNN architectures, CaffeNet \cite{jia2014caffe} (essentially AlexNet \cite{AlexNet}) and VGG16 \cite{VGG} proposed for an image classification task. The CaffeNet has 5 convolution layers, 2 fully connected layers and 1 softmax classification layer while VGG16 has 13 convolution layers, 2 fully connected layers and 1 softmax classification layer. Before the training of AOD, we first train a Fast R-CNN model using the above DCNN pre-trained on the ImageNet Classification task, following \cite{RCNN}. We then initialize all the convolution layers and 2 fully connected layers (fc6 and fc7 in Fig.~2 of the main manuscript) of the AOD by the corresponding layers in the trained Fast R-CNN model. 

\subsubsection{Other default settings}
Here we summarize some of the important parameters and design choices in our default network architecture. We set $T=3$ if not specifically mentioned. We set the standard deviations of the Gaussian random perturbation added to the generated glimpse representation to 0.2. The number of episodes generated from one sample is 8. Unlike a standard recurrent neural network, we have separate weights for a glimpse prediction layer at each time step. We empirically found this rendered a better performance.

\section{Experiments}

We evaluated the AOD network on 2007 and 2012 PASCAL VOC detection tasks~\cite{PASCAL_VOC} which contains 20 object classes. The performance of a detector is evaluated by the mean average precisions (mAP). For more details about the performance metric and evaluation protocol, please refer to~\cite{PASCAL_VOC}.

In this paper, we focused on validating the application of the attention mechanism for the visual object detection task. Hence, we only compared our results with those obtained by the baseline Fast R-CNN algorithm. Since the DCNN architecture employed had a significant impact on the final performance, we showed performance results separately based on the DCNN used. We also used the same proposal bounding boxes and the same pre-trained DCNN used in the Fast-RCNN work for a fair comparison.

We presented experiment results obtained under 4 different settings, which used different combinations of training and testing data as in~\cite{FastRCNN}. The VOC 2007 and VOC 2012 settings were the official settings, and the VOC 2007+2012 and VOC 2007++2012 were additional settings used for showing the effect of augmented training data. The training data in the VOC 2007+2012 consisted the training data from VOC 2007 and 2012 as well as the test data from VOC 2012. The training data in the VOC 2007++2012 consisted the training data from VOC 2007 and 2012 as well as the test data from VOC 2007. 

Table~\ref{table:mAP_results} summarizes mAP of the proposed methods as well as the baseline under different experiment settings. Full results including per-class Average Precision are shown in Table~\ref{table:VOC_2007_test_CaffeNet} to \ref{table:VOC_2012_test_VGG16_augmented}. When using CaffeNet~\cite{jia2014caffe} (essentially AlexNet~\cite{AlexNet}) as an underlying DCNN in the VOC 2007 setting, the AOD achieved an mAP of 58.1 when $T=3$ and an mAP of 57.8 when $T=2$, both outperforming the mAP of 57.1 obtained by the Fast R-CNN baseline. This validated the use of the attention mechanism for the visual object detection task. The proposed method improved the mAP value from 58.1 to 67.5 by using a stronger VGG16 net presented in \cite{VGG}. It again outperformed the Fast R-CNN baseline, which obtained 66.9 as using VGG16 net. The consistent improvements over the baseline as using a stronger DCNN suggested that the proposed algorithm could be applied with better DCNN to obtain a better performance. In the VOC 2012 setting, again, the proposed algorithm outperformed the baseline algorithm, improving the mAP value from 65.7 to 66.7. On larger datasets (2007+2012 and 2007++2012), the performance of the all the methods improved. The benefit of the attention mechanism was not downgraded with more training data.

\begin{figure*}[thb!]
  \centering
    \includegraphics[width=0.93\textwidth,natwidth=1546,natheight=948]{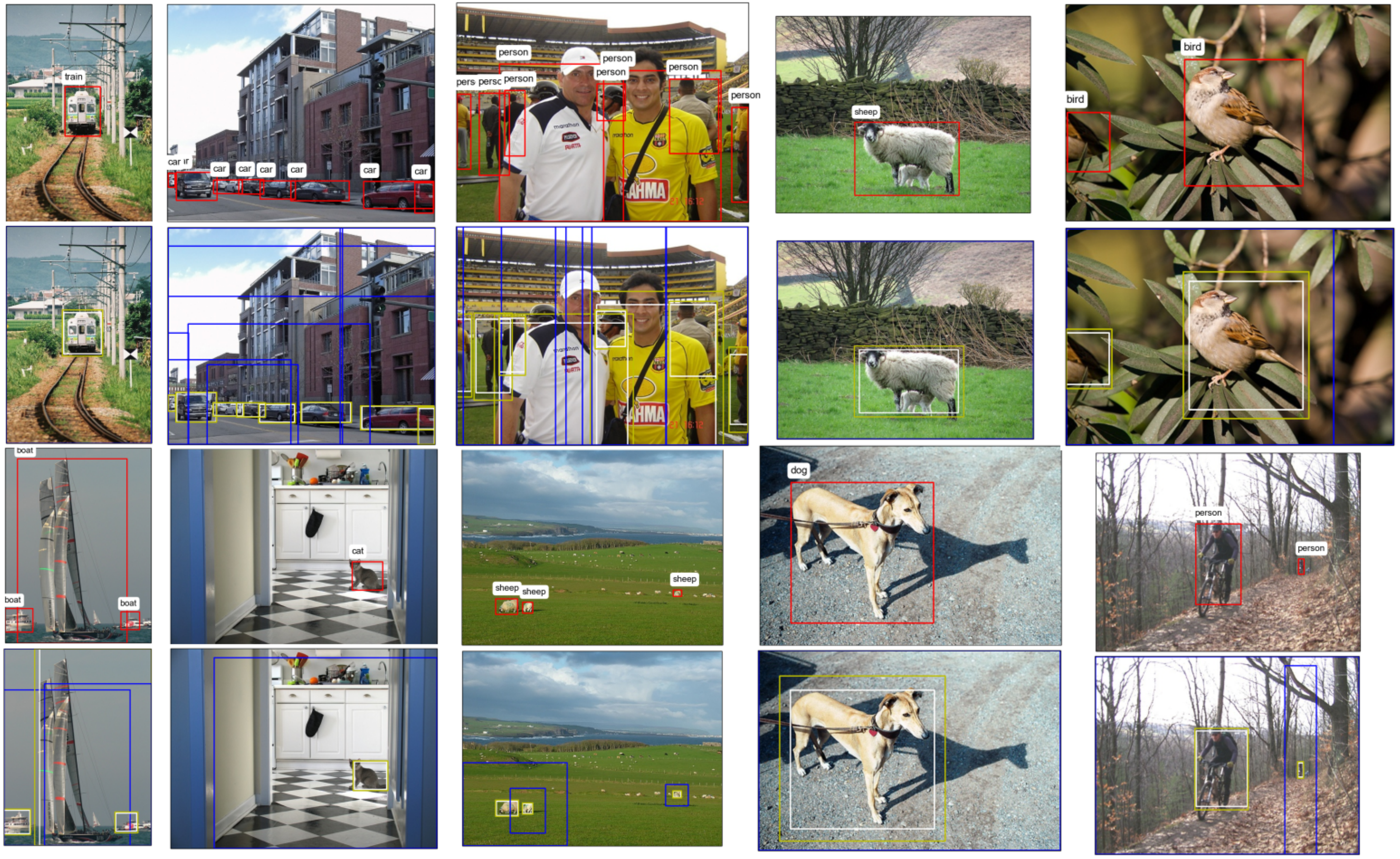}
    \caption{\small Example detection results. White, blue, yellow, and red bounding boxes represent object proposals, the first glimpses, the second glimpses and the final localization results, respectively.}
  \label{fig:detection_results}
\end{figure*}

In Fig.~\ref{fig:detection_results}, we show some example detection results using VGG16 under 2007+2012 setting. We first observed that AOD detected objects well. In the figure, we also visualized the learned glimpse. We found that the learned glimpse first tried to capture the context around the proposal bounding box and then looked at smaller regions.

\begin{table*}[t!]
\caption{Mean Average Precision of methods under various experimental settings}
\centering
\resizebox{\textwidth}{!}{%
    \begin{tabular}{|c||c|c|c|c|c|}
    \hline
     & CaffeNet & VGG16 & VGG16 & VGG16 & VGG16 \\
    \textbf{Methods} & VOC 2007 & VOC 2007 & VOC 2012 & VOC 2007+2012 & VOC 2007++2012 \\ \hline    
    Fast R-CNN & 57.1 & 66.9 & 65.7 & 70.0 & 68.4  \\ \hline 
    AOD T=2 & 57.8 & 67.1 & 66.5 & 71.1 & {\bf 69.5} \\ \hline
    AOD T=3 & {\bf 58.1} & {\bf 67.5} & {\bf 66.7} & {\bf 71.3} & 69.4 \\ \hline
    \end{tabular}}
\label{table:mAP_results}
\end{table*}

\subsection{Design Evaluation}\label{sec::design}

\begin{table}[h]
\centering
\caption{Effect of numbers of episodes generated from one sample in a mini-batch}
\begin{tabular}{|c||c|c|c|c|}
\hline
\# of episodes & 2 & 4 & 8 & 16 \\ \hline
mAP & 57.4 & 57.5 & 58.1 & 57.8 \\ \hline
\end{tabular}
\label{table:num_of_episodes}
\end{table}

\begin{table}[h]
\centering
\caption{The effect of the network architecture}
\begin{tabular}{|c|c|}
\hline
Network architecture & mAP \\ \hline
Stacked RNN with element-wise max & 58.1 \\ \hline
RNN with element-wise max & 57.4 \\ \hline
Stacked RNN without element-wise max & 57.0 \\ \hline
RNN without element-wise max & 57.2 \\ \hline
\end{tabular}
\label{table:network_architecture}
\end{table}

\begin{table}[h]
\centering
\caption{The effect of the reinforcement baseline methods}
\begin{tabular}{|c|c|}
\hline
Reinforcement baseline & mAP \\ \hline
Return normalization (ours) & 58.1 \\ \hline
Moving average & 57.8 \\ \hline
\end{tabular}
\label{table:baseline}
\end{table}

\begin{table}[h]
\centering
\caption{The effect of the choice between continuous return and discrete return}
\begin{tabular}{|c|c|}
\hline
Continuous return vs. discrete return & mAP \\ \hline
Continuous & 58.1 \\ \hline
Discrete & 57.8 \\ \hline
\end{tabular}
\label{table:continuous_or_discrete}
\end{table}

\begin{table}[h]
\centering
\caption{The effect of excluding background samples}
\begin{tabular}{|c|c|}
\hline
With background samples? & mAP \\ \hline
without background samples & 58.1 \\ \hline
with background samples & 57.6 \\ \hline
\end{tabular}
\label{table:excluding_bg_samples}
\end{table}

\begin{table}[h]
\centering
\caption{The effect of the glimpse representation}
\begin{tabular}{|c|c|}
\hline
Glimpse representation & mAP \\ \hline
x-shifting, y-shifting, x-scaling and y-scaling, & 58.1 \\ \hline
x-shifting, y-shifting & 57.3 \\ \hline
\end{tabular}
\label{table:fixed_glimpse}
\end{table}

We conducted extensive experimental evaluations to understand the impact of various design choices in the AOD network. The evaluations were conducted under the VOC 2007 setting with the CaffeNet.

\subsubsection{Number of episodes} We evaluated the impact of the number of episodes generated from one sample in a mini-batch (Table \ref{table:num_of_episodes}). As can be seen, the larger number of episodes tended to lead better performance. Since the computation time and the amount of memory also increased with the larger number of episodes, we picked 8 as the default number of episodes.

\subsubsection{Network architecture} We employed a stacked recurrent neural network, which had recurrent connections at both fc6 and fc7. We compared the default network architecture with a standard recurrent neural network, which had a recurrent connection only at fc7. In addition, we evaluated versions which directly performed the final classification and regression using the recurrent features at the last time step---without conducting the element-wise max (Elt-Max) operation. As shown in~\ref{table:network_architecture}, the stacked RNN with the element-wise max performed much better than the other architectures.

\subsubsection{Reinforcement baseline} We evaluated the effect of the reinforcement baselines. We compared our return normalization method presented in Sec.~\ref{sec:training} with the exponential moving average baseline. For the exponential moving average baseline, the result with the best smoothing parameter value obtained through a grid search was shown.

\subsubsection{Continuous return vs. discrete return} Our return was continuous (Eq.\eqref{eq:return}), ranging from 0 to 1. In \cite{mnih2014recurrent}, a discrete return was employed: a return was 1 if the highest scoring class was the ground-truth label and 0 otherwise. For validating the use of the continuous return, we adopted a similar discrete return computation where we assigned 1 if the highest scoring class was the ground-truth label AND an IoU between a predicted bounding box and the ground-truth bounding box was greater than or equal to the IoU threshold used in the evaluation. The results demonstrate the superiority of the continuous return over the discrete return (Table.~\ref{table:continuous_or_discrete}).

\subsubsection{Effect of excluding background samples} We evaluated the effect of excluding background samples from the REINFORCE algorithm. Since there was no ground-truth bounding boxes for background samples, we always set IOU in Eq.~\eqref{eq:return} to 1 for background samples. As can be seen in Table.~\ref{table:excluding_bg_samples}, excluding background samples yields a better performance.

\subsubsection{Glimpse representation} Our glimpse was represented as a four dimensional vector encoding x-shifting, y-shifting, x-scaling and y-scaling, enabling to generate an arbitrary glimpse bounding box. To evaluate the effect of different level of flexibility in representing glimpses, we conducted an experiment with a model employing two dimensional glimpse representation encoding only x-shifting and y-shifting (Table.~\ref{table:fixed_glimpse}). In other words, the glimpses generated for each proposal have one size. The experimental results clearly showed that allowing the network to produce arbitrary-shaped glimpse bounding boxes was important for achieving a good performance.

\begin{table*}[thb!]
\centering
\caption{Average Precision of methods using CaffeNet under the VOC 2007 setting}
\resizebox{\textwidth}{!}{%
    \begin{tabular}{c|cccccccccccccccccccc|c}
    \hline
    \textbf{Methods} & \textbf{aero} & \textbf{bike} & \textbf{bird} & \textbf{boat} & \textbf{bottle} & \textbf{bus} & \textbf{car} & \textbf{cat} & \textbf{chair} & \textbf{cow} & \textbf{table} & \textbf{dog} & \textbf{horse} & \textbf{mbike} & \textbf{person} & \textbf{plant} & \textbf{sheep} & \textbf{sofa} & \textbf{train} & \textbf{tv} & \textbf{mAP} \\ \hline
    Fast R-CNN & 66.4 & 71.6 & 53.8 & 43.3 & 24.7 & 69.2 & 69.7 & 71.5 & 31.1 & 63.4 & 59.8 & 62.2 & 73.1 & 65.9 & 57.0 & 26.0 & 52.0 & 56.4 & 67.8 & 57.7 & 57.1  \\ \hline 
    AOD T=2 & 66.4 & 72.9 & 51.1 & 44.4 & 24.8 & 66.5 & 71.2 & 72.5 & 30.2 & 66.3 & 63.0 & 65.0 & 74.1 & 68.5 & 58.3 & 25.5 & 50.5 & 55.8 & 71.2 & 56.9 & 57.8 \\ \hline
    AOD T=3 & 67.3 & 72.5 & 51.3 & 45.5 & 26.5 & 67.5 & 71.0 & 71.5 & 30.4 & 65.6 & 64.2 & 66.4 & 74.1 & 69.0 & 58.2 & 24.4 & 53.7 & 55.3 & 69.8 & 58.5 & 58.1 \\ \hline
    \end{tabular}}
\label{table:VOC_2007_test_CaffeNet}
\caption{Average Precision of methods using VGG16 under the VOC 2007 setting}
\resizebox{\textwidth}{!}{%
    \begin{tabular}{c|cccccccccccccccccccc|c}
    \hline
    \textbf{Methods} & \textbf{aero} & \textbf{bike} & \textbf{bird} & \textbf{boat} & \textbf{bottle} & \textbf{bus} & \textbf{car} & \textbf{cat} & \textbf{chair} & \textbf{cow} & \textbf{table} & \textbf{dog} & \textbf{horse} & \textbf{mbike} & \textbf{person} & \textbf{plant} & \textbf{sheep} & \textbf{sofa} & \textbf{train} & \textbf{tv} & \textbf{mAP} \\ \hline
    Fast R-CNN & 74.5 & 78.3 & 69.2 & 53.2 & 36.6 & 77.3 & 78.2 & 82.0 & 40.7 & 72.7 & 67.9 & 79.6 & 79.2 & 73.0 & 69.0 & 30.1 & 65.4 & 70.2 & 75.8 & 65.8 & 66.9  \\ \hline
    AOD T=2 & 74.9 & 78.1 & 64.9 & 51.3 & 40.8 & 80.1 & 78.5 & 80.6 & 42.9 & 74.1 & 68.4 & 78.2 & 79.9 &	76.5 & 69.4 & 32.1 & 64.4 & 67.1 & 74.7 & 65.5 & 67.1 \\ \hline
    AOD T=3 & 76.4 & 78.2 & 67.6 & 51.3 & 41.0 & 79.6 & 78.2 & 83.0 & 42.1 & 73.8 & 68.0 & 79.7 & 79.7 & 75.2 & 69.2 & 34.0 & 66.0 & 66.4 & 75.0 & 66.2 & 67.5 \\ \hline    
    \end{tabular}}
\label{table:VOC_2007_test_VGG16}
\caption{Average Precision of methods using VGG16 under the VOC 2012 setting}
\resizebox{\textwidth}{!}{%
    \begin{tabular}{c|cccccccccccccccccccc|c}
    \hline
    \textbf{Methods} & \textbf{aero} & \textbf{bike} & \textbf{bird} & \textbf{boat} & \textbf{bottle} & \textbf{bus} & \textbf{car} & \textbf{cat} & \textbf{chair} & \textbf{cow} & \textbf{table} & \textbf{dog} & \textbf{horse} & \textbf{mbike} & \textbf{person} & \textbf{plant} & \textbf{sheep} & \textbf{sofa} & \textbf{train} & \textbf{tv} & \textbf{mAP} \\ \hline
    Fast R-CNN & 80.3 & 74.7 & 66.9 & 46.9 & 37.7 & 73.9 & 68.6 & 87.7 & 41.7 & 71.1 & 51.1 & 86.0 & 77.8 & 79.8 & 69.8 & 32.1 & 65.5 & 63.8 & 76.4 & 61.7 & 65.7 \\ \hline
    AOD T=2 & 81.6 & 78.0 & 69.1 & 50.1 & 37.0 & 74.2 & 68.5 & 87.4 & 41.3 & 71.6 & 52.7 & 86.1 & 79.0 & 79.7 & 71.0 & 32.0 & 67.6 & 63.5 & 78.7 & 61.9 & 66.5 \\ \hline
    AOD T=3 & 82.5 & 77.6 & 69.7 & 50.0 & 37.4 & 74.2 & 68.7 & 87.0 & 41.8 & 71.4 & 52.8 & 85.7 & 78.9 &  79.6 & 70.9 & 32.8 & 67.6 & 63.9 & 78.9 & 61.8 & 66.7 \\ \hline
    \end{tabular}}
\label{table:VOC_2012_test_VGG16}
\caption{Average Precision of methods using VGG16 under the VOC 2007+2012 setting}
\resizebox{\textwidth}{!}{%
    \begin{tabular}{c|cccccccccccccccccccc|c}
    \hline
    \textbf{Methods} & \textbf{aero} & \textbf{bike} & \textbf{bird} & \textbf{boat} & \textbf{bottle} & \textbf{bus} & \textbf{car} & \textbf{cat} & \textbf{chair} & \textbf{cow} & \textbf{table} & \textbf{dog} & \textbf{horse} & \textbf{mbike} & \textbf{person} & \textbf{plant} & \textbf{sheep} & \textbf{sofa} & \textbf{train} & \textbf{tv} & \textbf{mAP} \\ \hline
    Fast R-CNN & 77.0 & 78.1 & 69.3 & 59.4 & 38.3 & 81.6 & 78.6 & 86.7 & 42.8 & 78.8 & 68.9 & 84.7 & 82.0 & 76.6 & 69.9 & 31.8 & 70.1 & 74.8 & 80.4 & 70.4 & 70.0 \\ \hline
    AOD T=2 & 77.6 & 78.6 & 70.1 & 59.7 & 38.2 & 83.3 & 79.3 & 87.6 & 48.3 & 78.9 & 71.8 & 83.5 & 84.0 & 78.8 & 71.7 & 33.1 & 73.3 & 74.3 & 80.0 & 70.2 & 71.1 \\ \hline
    AOD T=3 & 77.2 & 79.7 & 69.5 & 60.2 & 38.5 & 83.8 & 79.5 & 86.2 & 48.9 & 81.2 & 72.2 & 83.5 & 83.0  & 77.9 & 72.1 & 33.9 & 73.7 & 74.7 & 79.1 & 70.4 & 71.3 \\ \hline    
    \end{tabular}}
\label{table:VOC_2007_test_VGG16_augmented}
\caption{Average Precision of methods using VGG16 under VOC 2007++2012 setting}
\resizebox{\textwidth}{!}{%
    \begin{tabular}{c|cccccccccccccccccccc|c}
    \hline
    \textbf{Methods} & \textbf{aero} & \textbf{bike} & \textbf{bird} & \textbf{boat} & \textbf{bottle} & \textbf{bus} & \textbf{car} & \textbf{cat} & \textbf{chair} & \textbf{cow} & \textbf{table} & \textbf{dog} & \textbf{horse} & \textbf{mbike} & \textbf{person} & \textbf{plant} & \textbf{sheep} & \textbf{sofa} & \textbf{train} & \textbf{tv} & \textbf{mAP} \\ \hline
    Fast R-CNN & 82.3 & 78.4 & 70.8 & 52.3 & 38.7 & 77.8 & 71.6 & 89.3 & 44.2 & 73.0 & 55.0 & 87.5 & 80.5 & 80.8 & 72.0 & 35.1 & 68.3 & 65.7 & 80.4 & 64.2 & 68.4 \\ \hline
    AOD T=2 & 82.6 & 79.5 & 70.2 & 52.5 & 40.9 & 78.1 & 72.8 & 89.7 & 46.3 & 75.3 & 58.3 & 87.6 & 82.9 & 81.5 & 73.3 & 35.6 & 69.3 & 68.3 & 81.7 & 64.6 & 69.5 \\ \hline
    AOD T=3 & 82.2 & 79.6 & 70.5 & 52.7 & 40.5 & 78.5 & 72.8 & 88.9 & 45.8 & 75.6 & 57.7 & 87.5 & 82.5 & 80.9 & 73.6 & 35.3 & 69.6 & 67.5 & 80.8 & 64.6 & 69.4 \\ \hline
    \end{tabular}}
\label{table:VOC_2012_test_VGG16_augmented}
\end{table*}

\section{Conclusion}
We proposed an attentional network for visual object detection. It sequentially generated glimpse regions of various sizes and aspect ratios, extracted features from these regions, and made a final decision based on the information it had acquired. The key advantage of the proposed method was that the glimpses were adaptively generated in order to make more accurate decision. Since there were no ground truth annotations for glimpse locations and shapes, we trained the network using a reinforcement learning algorithm. The consistent performance improvement over the baseline method verified the benefit of incorporating the attention mechanism to the deep neural networks for the visual object detection task.

 


\bibliography{egbib,MyBib}

\begin{thebibliography}{10}
\providecommand{\url}[1]{#1}
\csname url@rmstyle\endcsname
\providecommand{\newblock}{\relax}
\providecommand{\bibinfo}[2]{#2}
\providecommand\BIBentrySTDinterwordspacing{\spaceskip=0pt\relax}
\providecommand\BIBentryALTinterwordstretchfactor{4}
\providecommand\BIBentryALTinterwordspacing{\spaceskip=\fontdimen2\font plus
\BIBentryALTinterwordstretchfactor\fontdimen3\font minus
  \fontdimen4\font\relax}
\providecommand\BIBforeignlanguage[2]{{%
\expandafter\ifx\csname l@#1\endcsname\relax
\typeout{** WARNING: IEEEtran.bst: No hyphenation pattern has been}%
\typeout{** loaded for the language `#1'. Using the pattern for}%
\typeout{** the default language instead.}%
\else
\language=\csname l@#1\endcsname
\fi
#2}}

\bibitem{ViolaJones}
P.~Viola and M.~J. Jones, ``Robust real-time face detection,'' \emph{IJCV},
  2004.

\bibitem{liu2016unsupervised}
M.-Y. Liu, A.~Mallya, O.~Tuzel, and X.~Chen, ``Unsupervised network pretraining
  via encoding human design,'' in \emph{2016 IEEE Winter Conference on
  Applications of Computer Vision (WACV)}.\hskip 1em plus 0.5em minus
  0.4em\relax IEEE, 2016, pp. 1--9.

\bibitem{RCNN}
R.~Girshick, J.~Donahue, T.~Darrell, and J.~Malik, ``Rich feature hierarchies
  for accurate object detection and semantic segmentation, tech report (v5),''
  \emph{arXiv:1311.2524v5}, 2014.

\bibitem{FastRCNN}
R.~Girshick, ``Fast r-cnn,'' in \emph{ICCV}, 2015.

\bibitem{mnih2014recurrent}
V.~Mnih, N.~Heess, A.~Graves, \emph{et~al.}, ``Recurrent models of visual
  attention,'' in \emph{NIPS}, 2014.

\bibitem{Chorowski2015}
J.~K. Chorowski, D.~Bahdanau, D.~Serdyuk, K.~Cho, and Y.~Bengio,
  ``Attention-based models for speech recognition,'' in \emph{NIPS}, 2015.

\bibitem{Bahdanau2015}
D.~Bahdanau, K.~Cho, and Y.~Bengio, ``Neural machine translation by jointly
  learning to align and translate,'' in \emph{ICLR}, 2015.

\bibitem{Sukhbaatar2015}
S.~Sukhbaatar, A.~Szlam, J.~Weston, and R.~Fergus, ``End-to-end memory
  networks,'' in \emph{NIPS}, 2015.

\bibitem{Caicedo2015}
J.~C. Caicedo and S.~Lazebnik, ``Active object localization with deep
  reinforcement learning,'' in \emph{ICCV}, 2015.

\bibitem{Yoo2015}
D.~Yoo, S.~Park, J.-Y. Lee, A.~S. Paek, and I.~S. Kweon, ``Attentionnet:
  Aggregating weak directions for accurate object detection,'' in \emph{ICCV},
  2015.

\bibitem{arXiv:1512.03385}
K.~He, X.~Zhang, S.~Ren, and J.~Sun, ``Deep residual learning for image
  recognition,'' \emph{CVPR}, 2016.

\bibitem{arXiv:1512.04412}
J.~Dai, K.~He, and J.~Sun, ``Instance-aware semantic segmentation via
  multi-task network cascades,'' \emph{CVPR}, 2016.

\bibitem{FasterRCNN}
S.~Ren, K.~He, R.~B. Girshick, and J.~Sun, ``Faster r-cnn: Towards real-time
  object detection with region proposal networks,'' in \emph{NIPS}, 2015.

\bibitem{arXiv:1506.02640}
J.~Redmon, S.~Divvala, R.~Girshick, and A.~Farhadi, ``You only look once:
  Unified, real-time object detection,'' \emph{CVPR}, 2016.

\bibitem{arXiv:1506.06981}
K.~Lenc and A.~Vedaldi, ``R-cnn minus r,'' \emph{arXiv:1506.06981}, 2015.

\bibitem{arXiv:1512.07729}
M.~Najibi, M.~Rastegari, and L.~S. Davis, ``G-cnn: an iterative grid based
  object detector,'' \emph{CVPR}, 2016.

\bibitem{MR-CNN}
S.~Gidaris and N.~Komodakis, ``Object detection via a multi-region \& semantic
  segmentation-aware cnn model,'' \emph{ICCV}, 2015.

\bibitem{chen2016small}
C.~Chen, M.-Y. Liu, O.~Tuzel, and J.~Xiao, ``R-cnn for small object
  detection,'' in \emph{Asian Conference on Computer Vision}, 2016.

\bibitem{arXiv:1512.04143}
S.~Bell, C.~L. Zitnick, K.~Bala, and R.~Girshick, ``Inside-outside net:
  Detecting objects in context with skip pooling and recurrent neural
  networks,'' \emph{CVPR}, 2016.

\bibitem{Sutton98}
R.~S. Sutton and A.~G. Barto, \emph{Reinforcement Learning: An Introduction
  (Adaptive Computation and Machine Learning)}.\hskip 1em plus 0.5em minus
  0.4em\relax {The MIT Press}, 1998.

\bibitem{SzepesvariBook10}
{\relax Cs}.~Szepesv\'ari, \emph{Algorithms for Reinforcement Learning}.\hskip
  1em plus 0.5em minus 0.4em\relax Morgan Claypool Publishers, 2010.

\bibitem{Williams1992}
R.~J. Williams, ``Simple statistical gradient-following algorithms for
  connectionist reinforcement learning,'' \emph{Machine Learning}, vol.~8, no.
  3-4, pp. 229--256, 1992.

\bibitem{DeisenrothNeumannPeters2013}
M.~P. Deisenroth, G.~Neumann, and J.~Peters, ``A survey on policy search for
  robotics.'' \emph{Foundations and Trends in Robotics}, vol.~2, no. 1-2, pp.
  1--142, 2013.

\bibitem{SuttonMcAleesterSinghMansour2000}
R.~S. Sutton, D.~McAllester, S.~Singh, and Y.~Mansour, ``Policy gradient
  methods for reinforcement learning with function approximation,'' in
  \emph{NIPS}, 2000.

\bibitem{58337}
P.~Werbos, ``Backpropagation through time: what it does and how to do it,''
  \emph{Proceedings of the IEEE}, vol.~78, no.~10, pp. 1550--1560, 1990.

\bibitem{jia2014caffe}
Y.~Jia, E.~Shelhamer, J.~Donahue, S.~Karayev, J.~Long, R.~Girshick,
  S.~Guadarrama, and T.~Darrell, ``Caffe: Convolutional architecture for fast
  feature embedding,'' \emph{ACM International Conference on Multimedia}, 2014.

\bibitem{AlexNet}
A.~Krizhevsky, I.~Sutskever, and G.~E. Hinton, ``Imagenet classification with
  deep convolutional neural networks,'' in \emph{NIPS}, 2012.

\bibitem{VGG}
K.~Simonyan and A.~Zisserman, ``Very deep convolutional networks for
  large-scale image recognition,'' in \emph{ICLR}, 2015.

\bibitem{PASCAL_VOC}
M.~Everingham, L.~Van~Gool, C.~Williams, J.~Winn, and A.~Zisserman, ``The
  pascal visual object classes (voc) challenge,'' \emph{IJCV}, 2010.

\end{thebibliography}
\bibliographystyle{IEEEtran}

\end{document}